\documentclass[lettersize,journal]{IEEEtran}
\usepackage{amsmath,amsfonts,amssymb}
\usepackage{algorithmic}
\usepackage{algorithm}
\usepackage{array}
\usepackage[caption=false,font=normalsize,labelfont=sf,textfont=sf]{subfig}
\usepackage{textcomp}
\usepackage{stfloats}
\usepackage{url}
\usepackage{verbatim}
\usepackage{graphicx}
\usepackage{cite}
\usepackage[hidelinks]{hyperref}
\usepackage{cleveref}
\usepackage[normalem]{ulem}
\usepackage{xspace}
\usepackage{xcolor}
\usepackage{booktabs}
\usepackage{multirow}

\newcommand{\eg}{\textit{e.g.}\xspace}
\begin{document}

\title{RefineSplat: Reconstruction-Guided Object-Centric Gaussian Splatting for Open-Vocabulary Querying}

\author{Yiren Lu, Disheng Liu, Sreya Srinidhi, Yunlai Zhou, Yiran Qiao, Chaoda Song, Tuo Liang, \\Jing Ma, Huan Wang, Yu Yin
\IEEEcompsocitemizethanks{%
  \IEEEcompsocthanksitem
  Yiren Lu, Disheng Liu, Sreya Srinidhi, Yunlai Zhou, Yiran Qiao, Chaoda Song, Tuo Liang, Jing Ma, Yu Yin are with the
  Department of Computer and Data Science, Case Western Reserve University,
  Cleveland, OH 44106 USA.
  E-mail: \{yiren.lu, yu.yin\}@case.edu.

  \IEEEcompsocthanksitem
  Huan Wang is with the Department of Artificial Intelligence, Westlake University, Hangzhou, Zhejiang 310024, P.R. China.
}
}

\maketitle

\begin{abstract}
Open-vocabulary 3D scene understanding aims to segment and retrieve arbitrary objects in 3D scenes using natural language queries. Most existing methods typically follow a \textit{splat-then-segment} paradigm, where 3D Gaussian Splatting is first reconstructed and then augmented with labels or language features for subsequent segmentation. However, this pipeline often suffers from view inconsistency, ambiguous object boundaries, and weak object-level correspondence, especially in complex static and dynamic scenes. Building on our prior \textit{Segment-then-Splat} framework, we present RefineSplat, a reconstruction-guided extension that retains its object-centric paradigm: it first tracks objects in 2D and then reconstructs each object with a dedicated set of 3D Gaussians, preserving explicit object-Gaussian correspondence for unified open-vocabulary segmentation and retrieval.

To improve robustness, we further close the loop between 2D tracking and 3D reconstruction. Since off-the-shelf video trackers can produce unreliable masks that propagate errors into object reconstruction, we introduce a reconstruction-guided feedback mechanism that refines object tracking using rendered masks and multi-view consistency. Moreover, instead of assigning each object a single language embedding, we propose a viewpoint-aware embedding assignment strategy, where each object maintains multiple view-clustered embeddings to capture view-dependent semantic cues. 
We also introduce \textit{Blender-OVS}, a synthetic benchmark of static and dynamic sequences with exact object meshes and point clouds, enabling direct evaluation of extracted geometry.
Experiments on static and dynamic benchmarks show improved rendered-view segmentation and lower object-level Chamfer distances, with competitive scene reconstruction quality.
\end{abstract}

\begin{IEEEkeywords}
Open-Vocabulary Scene Understanding, 3D Gaussian Splatting, Language Embedded Scene Representation, Language-guided Segmentation.
\end{IEEEkeywords}

\section{Introduction}
\begin{figure*}
    \centering
    \includegraphics[width=1\linewidth]{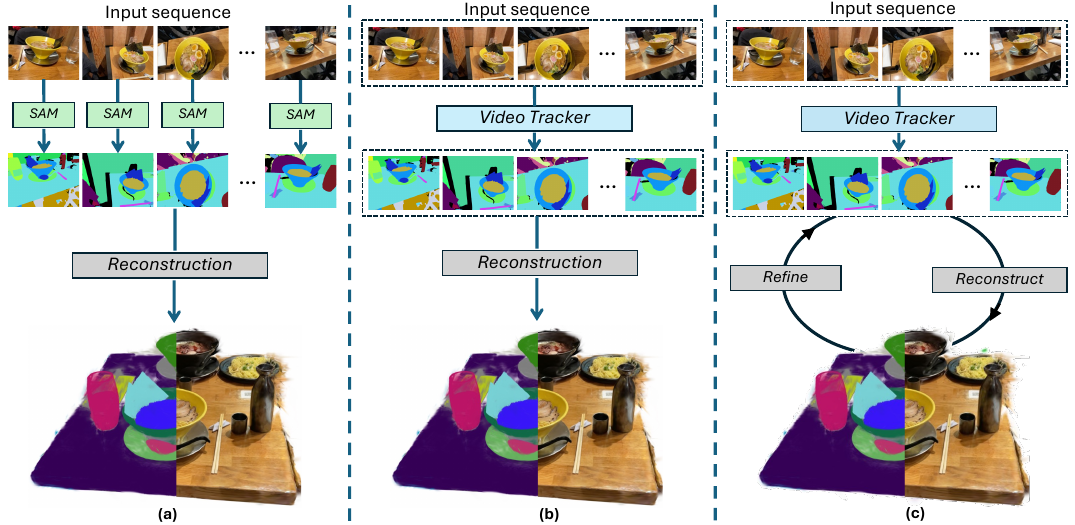}
    \caption{(a) \textbf{Independent per-view segmentation} produces object masks without persistent identities across views. (b) \textbf{Video tracking} propagates identities, but tracking failures can become direct supervision for the reconstructed object groups. (c) \textbf{\textit{RefineSplat}} closes the loop: track-level filtering uses an intermediate object-centric reconstruction to reject unreliable tracks before final reconstruction, while observation-level filtering selects suitable mask observations for object supervision during the later stage of final optimization. Mask colors denote object identities; the reconstructed scene is visualized with object IDs on the left and RGB appearance on the right.}
    \label{fig:teaser}
\end{figure*}
3D open-vocabulary querying is a key step toward language-driven interaction with 3D environments, allowing users to localize, segment, and retrieve objects without predefined category labels. Such a capability is fundamental for large-scale scene exploration, scene understanding, robotic navigation and manipulation, where free-form language serves as a natural interface between human intent and machine perception.

3D Gaussian Splatting (3DGS)~\cite{kerbl3Dgaussians} has recently become a compelling representation for this problem because it supports high-fidelity reconstruction, efficient optimization, and real-time differentiable rendering.
However, standard 3DGS is primarily a radiance representation: its anisotropic Gaussian primitives encode geometry and appearance, but not object identity or language-aligned semantics.
To enable open-vocabulary querying, recent methods~\cite{shi2024language,qin2024langsplat,zhou2024feature,gaussian_grouping} typically augment a reconstructed 3DGS scene with semantic labels or vision-language features.

We refer to this family as the \textit{splat-then-segment} paradigm because these methods first reconstruct the scene with 3DGS and subsequently assign semantic or identity embeddings to individual Gaussians for object segmentation and open-vocabulary querying.
Although it benefits from the reconstruction quality of 3DGS, it inherits a structural mismatch between scene-level radiance optimization and object-level language querying.
\textbf{First}, the Gaussians are optimized to explain the entire scene rather than to form object-level entities. 
Without explicit constraints on object boundaries and fine-grained geometry, individual Gaussians may overlap or span multiple objects, and object boundaries are only implicitly captured through appearance. This leads to object-level \textit{geometric ambiguity}. 
\textbf{Second}, language supervision is often imposed through alpha-blended feature rendering. 
This constrains the accumulated feature along a camera ray, but does not guarantee that each contributing Gaussian has an object-consistent semantic meaning.
The resulting \textit{semantic ambiguity} is especially pronounced around occlusion boundaries, thin structures, and regions where multiple objects project to nearby pixels. Consequently, a method may render a plausible 2D relevance map while lacking a clean 3D object representation that can be directly queried.

We reverse this pipeline and propose the \textbf{\textit{Segment-then-Splat}} framework~\cite{lu2026segment}. Instead of first reconstructing a scene-level Gaussian field and then searching for objects inside it, it first obtains multi-view object masks, assigns object identities to Gaussians, and reconstructs the scene as a composition of object-specific Gaussian sets.
Enforcing this object-level decomposition from the very beginning of reconstruction encourages Gaussians to align with object boundaries, and hence reduce geometric ambiguity.
Language features are then attached to each object rather than mixed along each viewing ray, reducing both geometric and semantic ambiguity.

Related studies, including COB-GS~\cite{zhang2025cobgs}, COS3D~\cite{zhu2025cos3d}, and ObjectGS~\cite{zhu2025objectgs}, also emphasize object-level structure in Gaussian representations.
These methods share the goal of aligning Gaussian representations with object identities and boundaries to support more reliable instance separation and object-level scene understanding.

However, for reconstruction pipelines that use initial 2D masks to define object-specific Gaussian groups, mask reliability remains a central limitation.
These masks are typically produced by off-the-shelf segmentation~\cite{kirillov2023segment} and video tracking models~\cite{ravi2025sam, cheng2023tracking}. Such 2D supervision may contain inaccurate boundaries, missed regions, identity switches, and temporal or cross-view inconsistencies.
Once these observations are used to initialize or supervise object-specific Gaussians, their errors can be propagated into the corresponding Gaussian groups and subsequently degrade both reconstruction and open-vocabulary querying, with no mechanism for recovery.

\looseness -1
This paper presents \textbf{\textit{RefineSplat}}, a reconstruction-guided extension of Segment-then-Splat that closes the loop between 2D supervision and 3D object reconstruction (Fig.~\ref{fig:teaser}(c)).
Our key insight is that an intermediate object-specific reconstruction consolidates evidence across multiple views and can thus serve as a geometric consistency signal for assessing the reliability of its 2D supervision.
We render reconstructed objects into the training views and compare their support with the tracked masks to reject globally unreliable tracks (track-level filtering) and select suitable observations for object-level supervision (observation-level filtering).
The retained masks then provide cleaner supervision for the final object-specific Gaussian optimization, preventing tracking errors from propagating into 3D.

RefineSplat further revisits how language information should be attached to object-centric Gaussians.
Prior object-centric methods, including our conference version, collapse all observations of an object into a single globally averaged embedding, which suppresses view-dependent semantics (\eg, the face of a clock versus its back).
We instead introduce a viewpoint-aware embedding assignment strategy.
By grouping object observations according to viewpoint and maintaining multiple view-clustered embeddings for each object, the proposed strategy preserves view-dependent semantic cues and enables more discriminative open-vocabulary querying.

Finally, we address an evaluation gap in open-vocabulary 3D segmentation.
Existing benchmarks such as LERF~\cite{kerr2023lerf}, Mip-NeRF360~\cite{barron2022mip}, and HyperNeRF~\cite{park2021hypernerf} rely on 2D mask annotations and evaluate segmentation in the rendered image space. A method can score well on rendered views while its underlying 3D object geometry is incomplete or entangled with neighboring objects.
We therefore introduce \textbf{\textit{Blender-OVS}}, a benchmark covering both static and dynamic scenes with exact 3D-level ground truth, including per-object meshes and point clouds.
Together with direct 3D evaluation, Blender-OVS complements rendered-view metrics and enables a more faithful assessment of 3D open-vocabulary segmentation.

This paper is an extended version of our NeurIPS 2025 work, ``Segment then Splat: Unified 3D Open-Vocabulary Segmentation via Gaussian Splatting''~\cite{lu2026segment}.
Building on the explicit object-Gaussian correspondence established by the original framework, this work makes the following new contributions:
\begin{enumerate}
    \item \textbf{Reconstruction-guided supervision refinement.} We introduce \textit{RefineSplat}, a reconstruction-guided extension of \textit{Segment-then-Splat} that uses intermediate object-specific reconstructions to identify unreliable 2D supervision, thereby mitigating the propagation of tracking errors into 3D reconstruction.

    \item \textbf{Viewpoint-aware open-vocabulary querying.} We represent each object with multiple view-clustered language embeddings rather than one globally averaged descriptor, preserving view-dependent semantic cues.

    \item \textbf{Blender-OVS benchmark and direct 3D evaluation.} We introduce \textit{Blender-OVS}, a 3D open-vocabulary segmentation benchmark with exact per-object meshes and point clouds for both static and dynamic scenes, together with an evaluation protocol that measures the geometry of queried objects directly in 3D, beyond 2D views.
\end{enumerate}

Experiments on static and dynamic benchmarks show improved rendered-view segmentation and lower average Chamfer distances for queried 3D objects compared with the evaluated baselines, while maintaining competitive scene reconstruction quality.

\section{Related Work}
\subsection{Gaussian Scene Representations}

Neural radiance fields (NeRF)~\cite{mildenhall2020nerf} have driven substantial progress in novel-view synthesis by representing a scene as a continuous volumetric field and rendering it with differentiable volume integration.
However, their dense ray sampling and repeated network evaluation make training and rendering costly, especially when additional semantic or language fields are optimized on top of them.
3D Gaussian Splatting (3DGS)~\cite{kerbl3Dgaussians} replaces implicit volumetric sampling with explicit anisotropic Gaussian primitives and fast differentiable rasterization, achieving high-fidelity reconstruction with real-time rendering.
This efficiency has made 3DGS a strong substrate for 3D scene understanding beyond view synthesis.

Dynamic 3DGS and 4D Gaussian representations further extend this explicit representation to time-varying scenes~\cite{yang2023gs4d,yang2024deformable,Wu_2024_CVPR}.
These methods typically combine a canonical Gaussian representation with time-dependent deformation, enabling reconstruction of moving objects and changing visibility.
Together, static and dynamic Gaussian representations provide the reconstruction backbone for open-vocabulary querying, but their original formulations model appearance and geometry rather than object identity or language semantics.

\subsection{Language-Embedded Scene Representations}

Open-vocabulary 3D scene understanding requires associating 3D structures with free-form language queries. 
Early approaches build upon neural radiance fields \cite{kobayashi2022decomposing, kerr2023lerf, siddiqui2023panoptic, engelmann2024opennerf, tschernezki2022neural}, distilling features from vision-language or segmentation foundation models into continuous 3D fields. 
These methods demonstrate the potential of combining neural scene representations with open-vocabulary semantics, but they often inherit the high computational cost of implicit neural fields, especially during optimization and rendering.

Gaussian-based methods associate language with explicit primitives or auxiliary feature fields.
LEGaussians~\cite{shi2024language} and LangSplat~\cite{qin2024langsplat} compress language features for efficient scene-specific learning, while Feature 3DGS~\cite{zhou2024feature} distills foundation-model features through multi-view rendering.
FMGS~\cite{zuo2025fmgs} combines Gaussian Splatting with multi-resolution hash encodings and pixel-aligned feature supervision.
GAGS~\cite{peng2026gags} addresses inconsistent feature granularity, and LangSplatV2~\cite{li2025langsplatv2} learns sparse coefficients over a shared dictionary to render language features efficiently.
Beyond scene-specific optimization, OVGaussian~\cite{chen2024ovgaussian} trains a cross-scene Gaussian segmentation network using cross-modal consistency learning.

A complementary line lifts features onto reconstructed Gaussian scenes without additional semantic-feature training.
Occam's LGS~\cite{cheng2025occamlgs} aggregates multi-view language features using alpha-blending weights and filters noisy Gaussians.
LUDVIG~\cite{marrie2025ludvig} uplifts visual features through weighted aggregation and refines them with graph diffusion informed by 3D geometry and feature similarities.
These approaches avoid iterative feature-field fitting, but feature aggregation may also introduce semantic ambiguity.

For dynamic scenes, language-embedded Gaussian representations further model time-varying semantics.
DGD~\cite{labe2024dgd} distills 2D semantic features into dynamic 3D Gaussians, enabling prompt-based segmentation and tracking over time.
4D LangSplat~\cite{li20254dlangsplat4dlanguage} uses object-wise video captions generated by multimodal large language models to construct a 4D language field for time-sensitive and time-agnostic queries.
LangField4D~\cite{xu2026langfield4d} introduces identity-adaptive Gaussian grouping to model time-varying object affiliations and learns continuous spatio-temporal language features.

\subsection{Instance Grouping and Object-Level Geometry}

Instance-aware methods explicitly address the grouping of Gaussian primitives.
Gaussian-Grouping~\cite{gaussian_grouping} learns object identity encodings from tracked masks, while OpenGaussian~\cite{wu2024opengaussian} learns point-level features and associates them with language.
GAGA~\cite{lyu2026gaga} associates masks across views through a 3D-aware memory bank, reducing dependence on continuous camera trajectories.
SAGA~\cite{cen2025segment} learns scale-gated affinity features for multi-granularity segmentation from visual prompts.
ObjectGS~\cite{zhu2025objectgs} couples object-aware reconstruction and scene understanding.
SuperGSeg~\cite{liang2026supergseg} organizes Gaussians into sparse Super-Gaussians carrying language features and supporting hierarchical segmentation.
COS3D~\cite{zhu2025cos3d} connects instance and language fields through collaborative training and adaptive query refinement.

Other methods modify Gaussian geometry to improve object boundaries.
SAGD~\cite{hu2026sagd} introduces a training-free segmentation pipeline with decomposition of boundary Gaussians.
COB-GS~\cite{zhang2025cobgs} uses semantic-gradient statistics for boundary-adaptive splitting and subsequently restores appearance.
BEA-GS~\cite{mazzucchelli2026beags} optimizes both visible and occluded Gaussian geometry for precise object extraction.
These studies highlight that semantic assignment and geometric separation are distinct requirements.
Mask-dependent pipelines remain sensitive to errors in their 2D supervision, with motion and changing visibility introducing additional inconsistencies.
Following Segment-then-Splat~\cite{lu2026segment}, RefineSplat assigns object identities before reconstruction and optimizes object-specific Gaussian groups.
Its reconstruction-guided feedback additionally addresses supervision reliability in both static and dynamic scenes by filtering inconsistent tracks and object--view observations before or during final optimization.

\section{Method}
\begin{figure*}[t]
    \centering
    \includegraphics[width=1\linewidth]{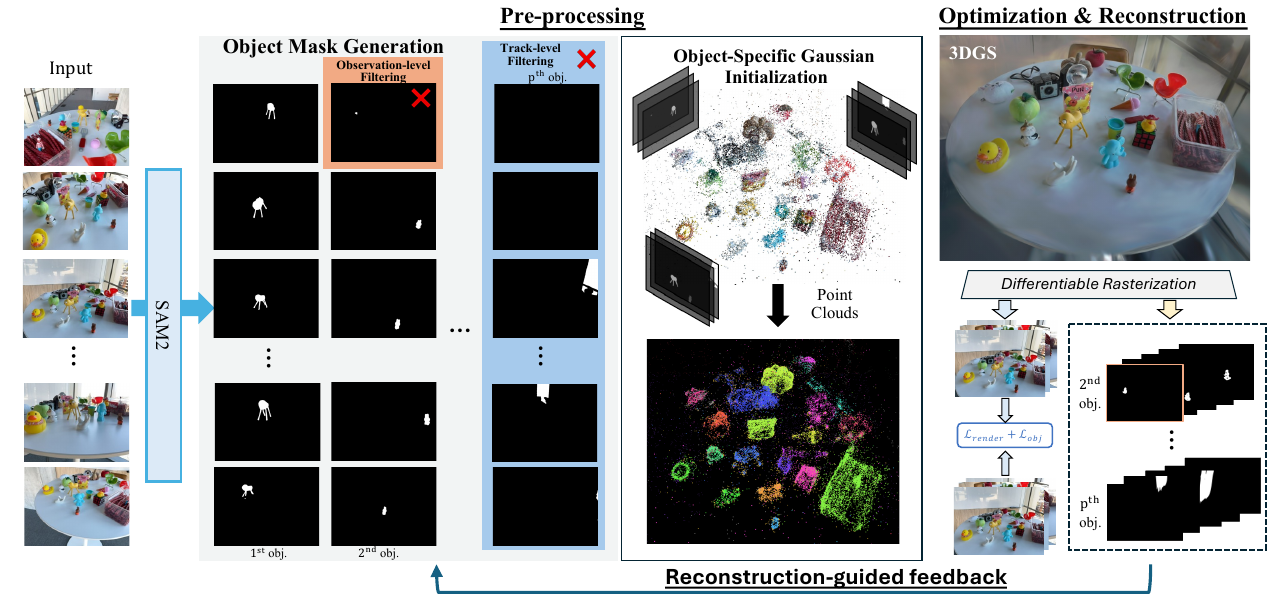}
    \caption{\textbf{Overview of the reconstruction-guided object-centric pipeline}. SAM2 propagates class-agnostic object masks through the calibrated input sequence. Multi-view votes assign persistent object identities to the initial Gaussians, and object-specific rendering losses preserve these identities during densification and reconstruction. RefineSplat first optimizes an intermediate model, renders occlusion-aware object-membership maps, and removes tracks whose disagreement with the tracked masks is frequent and severe across valid views (track-level filtering). The retained tracks are then reinitialized for final reconstruction, during which observation-level filtering selects suitable mask observations for object-level supervision in the later optimization stage.}
    \label{fig:framework}
\end{figure*}
We introduce \textbf{\textit{RefineSplat}}, an object-centric framework for 3D open-vocabulary segmentation and retrieval in static and dynamic 3D scenes based on Gaussian Splatting. 
An overview of our reconstruction-guided pipeline is shown in \cref{fig:framework}.
Following \textit{Segment-then-Splat}~\cite{lu2026segment}, our framework decomposes the scene into object-level components prior to reconstruction, reversing the conventional splat-then-segment pipelines and enforcing an explicit object-Gaussian correspondence.

Given multi-view object masks produced by an off-the-shelf video tracking model (\cref{subsec:object_tracking}), we assign each initialized Gaussian an object ID at each mask granularity, forming object-specific Gaussian sets at both levels (\cref{subsec:gs_initialization}). 
After that we perform object-level reconstruction (\cref{subsec:optimization}), to better preserve Gaussian-object correspondence and improve object-level geometry.

On top of this backbone, RefineSplat contributes two new components.
First, to prevent tracking errors from propagating into the 3D representation, we introduce a reconstruction-guided supervision filtering mechanism (\cref{subsec:mask_refinement}) that uses intermediate object-level reconstructions as feedback to reject unreliable object tracks and individual mask observations.
Second, after reconstruction, we propose a viewpoint-aware embedding assignment strategy (\cref{subsec:viewpoint_aware_embedding}) that equips each object with multiple view-clustered language embeddings, capturing view-dependent semantic cues for open-vocabulary querying (\cref{subsec:open_vocab_query}).

\subsection{Preliminary: 3D and 4D Gaussian Splatting}
\label{subsec:preliminaries}

For a static scene, 3D Gaussian Splatting (3DGS) \cite{kerbl3Dgaussians} represents a scene as a set of anisotropic Gaussian primitives 
$\mathcal{G}=\{G_i\}_{i=1}^{N}$. 
Each Gaussian $G_i$ is parameterized by a 3D center $\boldsymbol{\mu}_i \in \mathbb{R}^3$, a covariance matrix $\boldsymbol{\Sigma}_i \in \mathbb{R}^{3 \times 3}$, an opacity $\alpha_i$, and view-dependent color attributes, typically modeled by spherical harmonics (SH). 

A key advantage of 3DGS is its real-time differentiable rendering capability. 
Given a scene represented by 3D Gaussians and a target camera pose, each 3D Gaussian is projected onto the image plane as a 2D Gaussian, whose projected mean and covariance determine its spatial support in the rendered view. 
The final pixel color is computed through differentiable alpha compositing over depth-sorted Gaussians:
\begin{equation}
\mathbf{C}(u) = \sum_{i \in \mathcal{N}(u)} T_i \alpha_i(u) \mathbf{c}_i,
\quad
T_i = \prod_{j<i} \left(1-\alpha_j(u)\right),
\label{eq:alpha_compositing}
\end{equation}
where $u$ denotes a pixel location, $\mathcal{N}(u)$ is the set of Gaussians contributing to that pixel, $\mathbf{c}_i$ is the color of Gaussian $G_i$, and $T_i$ is the accumulated transmittance before the $i$-th Gaussian. 
This rasterization process is fully differentiable, allowing gradients to flow from rendered images back to Gaussian attributes through projection, splatting, and alpha compositing.

For dynamic sequences, we adopt the 4D Gaussian Splatting (4DGS) backbone~\cite{Wu_2024_CVPR}, which combines canonical 3D Gaussians with a multi-resolution HexPlane feature field and a lightweight deformation decoder.
The feature field comprises six learnable planes over the coordinate pairs $\mathcal{A}=\{xy,xz,xt,yz,yt,zt\}$ at each resolution level.
Given the canonical center $\boldsymbol{\mu}_i$ and timestamp $t$, we bilinearly sample each plane, multiply the sampled features element-wise within each level, and concatenate the results across levels:
\begin{equation}
\mathbf{h}_i(t) =
\bigoplus_{\ell=1}^{L}
\left(
\bigodot_{ab\in\mathcal{A}}
\mathbf{P}_{ab}^{(\ell)}\!\left(\pi_{ab}(\tilde{\boldsymbol{\mu}}_i,t)\right)
\right),
\label{eq:hexplane_features}
\end{equation}
where $\tilde{\boldsymbol{\mu}}_i$ is the center normalized to the initial point cloud's bounding box, $\pi_{ab}$ selects the corresponding coordinate pair, $\mathbf{P}_{ab}^{(\ell)}$ denotes bilinear feature sampling on that plane, and $\bigodot$ and $\bigoplus$ denote element-wise multiplication and concatenation, respectively.
A shared MLP with separate output heads decodes these features into position, quaternion, and scale offsets:
\begin{equation}
(\Delta\boldsymbol{\mu}_i(t),\Delta\mathbf{r}_i(t),
\Delta\mathbf{s}_i(t))
=\mathcal{F}_{\phi}\!\left(\mathbf{h}_i(t)\right),
\label{eq:deformation}
\end{equation}
The offsets are added to the canonical center $\boldsymbol{\mu}_i$, rotation quaternion $\mathbf{r}_i$, and scale vector $\mathbf{s}_i$ before rendering with \cref{eq:alpha_compositing}.
Opacity and SH coefficients are optimized as time-independent attributes, and each Gaussian retains its object identity across timestamps.
The same time-conditioned geometry is used for full-scene rendering, object-level supervision, and reconstruction-guided feedback.

\subsection{Class-Agnostic Object Tracking}
\label{subsec:object_tracking}

Given an image sequence $\{I_t\}_{t=0}^{n}$, we obtain masks $M_t^o$ for object $o$ in each frame/view $t$ at multiple granularity levels. 
To initialize object candidates, we apply Segment Anything (SAM)~\cite{kirillov2023segment} to the first frame $I_0$ to produce a set of anything masks that capture objects and object parts at different scales. 
These masks are then used as object prompts for SAM2~\cite{ravi2025sam} to produce temporally associated multi-view masks.

\subsubsection{\textbf{New Object Discovery}}
Objects absent from the initial frame cannot be recovered by forward propagation alone. We therefore periodically check the tracking results every $\Delta t$ frames.
At each check, we compare the mask-covered image area with that of the current reference frame.
A significant decrease indicates that some visible regions are not explained by existing tracks, suggesting new incoming objects.
We then re-apply SAM to the current frame and identify the mask regions with little overlap with existing tracks as new object prompts.
The updated prompts are provided to SAM2 for subsequent tracking.
The discovery frame becomes the reference for subsequent coverage checks.

\subsubsection{\textbf{Conflict Resolution}}
Since object masks are propagated independently, different instance tracks may occasionally overlap when tracking error occurs. 
To resolve such conflicts, we introduce an overlap resolution step. 
Let $M_t^o \in \{0,1\}^{H \times W}$ denote the binary mask of object $o$ at frame $t$. 
For two distinct objects $o$ and $o'$, we define the overlapping region of their masks as
\begin{equation}
J_t^{o,o'} = M_t^o \cap M_t^{o'} .
\end{equation}
A pair is considered conflicting if the overlap is non-trivial and satisfies either a large pairwise IoU or a high containment ratio:
\begin{equation}
\frac{|M_t^o \cap M_t^{o'}|}{|M_t^o \cup M_t^{o'}|} > \tau_{\mathrm{iou}}
\quad \text{or} \quad
\max \left(
\frac{|J_t^{o,o'}|}{|M_t^o|},
\frac{|J_t^{o,o'}|}{|M_t^{o'}|}
\right) > \tau_c ,
\end{equation}
where $\tau_{\mathrm{iou}}$ and $\tau_c$ are predefined thresholds. 
Very small overlaps are ignored in practice.

For each conflicting mask, we estimate a reliability score based on its temporal stability:
\begin{equation}
S_t^o =
\lambda_{\mathrm{IoU}} C_{\mathrm{IoU}}^o
+
\lambda_{\mathrm{shape}} C_{\mathrm{shape}}^o
+
\lambda_{\mathrm{motion}} C_{\mathrm{motion}}^o ,
\end{equation}
where $C_{\mathrm{IoU}}^o$ measures mask consistency with the previous frame, $C_{\mathrm{shape}}^o$ measures area stability, and $C_{\mathrm{motion}}^o$ measures centroid motion consistency. 
Specifically, these terms are defined as
\begin{equation}
\begin{gathered}
C_{\mathrm{IoU}}^o =
\frac{|M_t^o \cap M_{t-1}^o|}{|M_t^o \cup M_{t-1}^o|},
\quad
C_{\mathrm{shape}}^o =
\frac{\bar{A}_{t-1}^o}
{\bar{A}_{t-1}^o + |A_t^o - \bar{A}_{t-1}^o|}, \\
C_{\mathrm{motion}}^o =
\exp\left(
-\frac{\|p_t^o - p_{t-1}^o\|_2}{0.1\sqrt{H^2 + W^2}}
\right).
\end{gathered}
\end{equation}
Here, $A_t^o=|M_t^o|$ denotes the mask area, $\bar{A}_{t-1}^o$ is the recent average area of object $o$, $p_t^o$ is the centroid of $M_t^o$, and $H$ and $W$ denote the image height and width, respectively.

For each conflicting pair $(o,o')$, we keep the overlapping pixels for the mask with the higher reliability score and remove them from the other mask.
This assigns each pixel to at most one instance while preserving the mask track that is more temporally stable and spatially consistent.

\subsection{Object-centric Reconstruction} \label{subsec:reconstruction}
\subsubsection{\textbf{Object-specific Gaussian Initialization}}
\label{subsec:gs_initialization}

To initialize object-specific Gaussians, we assign object labels to sparse COLMAP~\cite{schoenberger2016sfm, schoenberger2016mvs} points using multi-view mask observations.
Let $X_k$ denote a 3D point reconstructed by COLMAP, and let $\mathcal{V}(X_k)$ be the set of training views in which $X_k$ is observed.
For each view $t \in \mathcal{V}(X_k)$, COLMAP provides the corresponding 2D observation $u_{k,t} \in \mathbb{R}^2$.
We perform voting independently at each granularity $g\in\{\mathrm{middle},\mathrm{large}\}$.
Let $\mathcal{O}_g$ denote the object labels at granularity $g$ and $M_t^{o,g}$ the corresponding mask of object $o$ in view $t$.
The vote count for assigning point $X_k$ to object $o$ at granularity $g$ is
\begin{equation}
n_k^{o,g} = \sum_{t \in \mathcal{V}(X_k)}
\mathbb{I}\left[u_{k,t} \in M_t^{o,g}\right],
\end{equation}
where $\mathbb{I}[\cdot]$ denotes the indicator function.
The object label of $X_k$ is determined separately within each granularity:
\begin{equation}
o_k^g = \arg\max_{o\in\mathcal{O}_g} n_k^{o,g}.
\end{equation}
If no mask receives a vote at a given granularity, the point is assigned the background label at that level.
Each COLMAP point is initialized as a single Gaussian carrying both labels $(o_k^{\mathrm{middle}},o_k^{\mathrm{large}})$.
In the following, we omit the granularity superscript for brevity and use $o$ to identify a group by its granularity and object label.

\subsubsection{\textbf{Optimization \& Reconstruction}}
\label{subsec:optimization}

After initializing object-labeled Gaussians from sparse COLMAP points, we optimize them to reconstruct the full scene while preserving object correspondence.
For the full-scene reconstruction, we adopt the standard photometric rendering loss:
\begin{equation}
    \mathcal{L}_{\text{render}} =
    (1-\lambda_r)\mathcal{L}_1(\hat{I}_t, I_t)
    + \lambda_r \mathcal{L}_{\text{DSSIM}}(\hat{I}_t, I_t),
\end{equation}
where $I_t$ and $\hat{I}_t$ denote the input image and the rendered image at view $t$, respectively.

To further regularize object-level reconstruction, we render each object-specific Gaussian set independently.
Let $\hat{I}_t^o$ and $\hat{A}_t^o$ denote the rendered image and alpha map of object $o$ at view $t$, and let $M_t^o$ be the corresponding object mask obtained from \cref{subsec:object_tracking}.
Let $\Omega$ denote the image pixel grid and $u\in\Omega$ a pixel location.
We define the object RGB loss as
\begin{equation}
    \mathcal{L}_{\text{rgb}}^o =
    \frac{\sum_{u\in\Omega} M_t^o(u)
    \left\|\hat{I}_t^o(u)-I_t(u)\right\|_1}
    {3\sum_{u\in\Omega} M_t^o(u)},
\end{equation}
The denominator normalizes over the RGB elements inside the non-empty object mask.
This loss constrains the rendered appearance within the associated object region, while $\mathcal{L}_{\text{render}}$ supervises the full-scene image.
In addition, we introduce an alpha consistency loss to explicitly constrain the rendered opacity of each object:
\begin{equation}
    \mathcal{L}_{\alpha}^o =
    \mathcal{L}_1(\hat{A}_t^o, M_t^o).
\end{equation}
The alpha loss encourages the object-specific Gaussians to be opaque inside the object mask and transparent outside it, which helps suppress leakage across neighboring objects and background regions.
The object-level loss for object $o$ is therefore defined as
\begin{equation}
    \mathcal{L}_{\text{obj}}^o = 
    \mathcal{L}_{\text{rgb}}^o
    + \lambda_{\alpha}\mathcal{L}_{\alpha}^o,
\end{equation}
where $\lambda_{\alpha}$ balances the RGB and alpha constraints.
The final optimization objective is
\begin{equation}
    \mathcal{L} =
    \mathcal{L}_{\text{render}}
    + \lambda_{\text{obj}}\mathcal{L}_{\text{obj}}
    + \mathcal{L}_{\text{reg}},
\end{equation}
where $\lambda_{\text{obj}}$ controls the contribution of object-level loss and $\mathcal{L}_{\text{reg}}=0$ for static scenes.
For dynamic sequences, we additionally regularize the HexPlane field following 4DGS~\cite{Wu_2024_CVPR}:
\begin{equation}
\mathcal{L}_{\text{reg}} =
\lambda_{\mathrm{sp}}\mathcal{R}_{\mathrm{sp}}
+ \lambda_{\mathrm{tm}}\mathcal{R}_{\mathrm{tm}}
+ \lambda_{\mathrm{l1}}\mathcal{R}_{\mathrm{l1}},
\end{equation}
where $\mathcal{R}_{\mathrm{sp}}$ penalizes variations in the spatial planes, $\mathcal{R}_{\mathrm{tm}}$ penalizes second-order temporal differences in the space--time planes, and $\mathcal{R}_{\mathrm{l1}}$ is an $L_1$ penalty pulling the space--time plane values toward one.

During optimization, densification, cloning, and pruning are performed independently within each object-specific Gaussian set.
This prevents newly created Gaussians from changing their object identities and preserves the Gaussian-object correspondence throughout reconstruction.
We also keep a minimum number of Gaussians for each object set to avoid the case where all Gaussians of an object are removed during pruning.

\subsubsection{\textbf{Efficient Object-Level Supervision}}
Applying object-level supervision to every object at every iteration would require rendering all object-specific Gaussian sets, which becomes expensive when a scene contains many objects.
Therefore, instead of optimizing all objects simultaneously, we randomly sample a subset of $m$ objects at each iteration and apply the object-level loss only to the sampled objects.
This strategy reduces additional rendering cost while maintaining sufficient object-level regularization.

\subsubsection{\textbf{Multi-Granularity Optimization}}
Our masks contain two levels of granularity: middle-level objects and large-level objects.
Middle-level masks capture finer parts or sub-regions, while large-level masks provide broader object-level constraints.
The same Gaussian participates in the groups specified by its two independently assigned labels; the two levels share the underlying Gaussian representation.
We incorporate supervision at both levels during reconstruction, applying the object RGB and alpha losses to sampled Gaussian groups at the corresponding granularity.
This multi-granularity supervision supports both detailed object structure and larger semantic regions while preserving the associated object identities.

\subsection{Reconstruction-guided Supervision Filtering}
\label{subsec:mask_refinement}

We refine supervision at two scopes: removing globally unreliable tracks and selecting informative observations for the retained objects.
These are implemented through track-level and observation-level reconstruction feedback, respectively.

\subsubsection{\textbf{Track-level Filtering}}
We first train an initial object-centric Gaussian representation using the tracked masks.
For each object $o$, we render an occlusion-aware object-membership map into every training view by assigning a value of $1$ to its Gaussians and $0$ to all other Gaussians while retaining the original depth ordering and alpha compositing.
This produces a soft visibility map $Z_t^o(u) \in [0,1]$, which we threshold to obtain a reconstruction-induced mask:
\begin{equation}
\hat{M}_t^o(u) =
\mathbb{I}\left[Z_t^o(u) > \tau_v\right],
\end{equation}
where $u\in\Omega$ is a pixel location and $\tau_v$ is the visibility threshold.
We compare $\hat{M}_t^o$ with the tracked mask $M_t^o$ using
\begin{equation}
\rho_t^o =
\frac{
|\hat{M}_t^o \cap M_t^o|
}{
|\hat{M}_t^o \cup M_t^o|
}.
\end{equation}
Views in which either mask is too small or their areas are severely imbalanced are excluded from this comparison.
Let $\mathcal{V}_o$ denote the remaining valid views and $\mathcal{D}_o=\{t\in\mathcal{V}_o\mid\rho_t^o<\tau_{\mathrm{bad}}\}$ denote the views that are inconsistent with the reconstructed object.

We aggregate the feedback over views using the inconsistent-view ratio and median IoU:
\begin{equation}
r_{\mathrm{bad}}^o =
\frac{|\mathcal{D}_o|}{|\mathcal{V}_o|},
\qquad
\tilde{\rho}^{o} =
\underset{t\in\mathcal{V}_o}{\mathrm{median}}\left(\rho_t^o\right).
\end{equation}
Given a minimum number of valid comparisons, we remove the entire object track if
\begin{equation}
r_{\mathrm{bad}}^o \ge \tau_r
\quad \text{and} \quad
\tilde{\rho}^{o} \le \tau_m,
\end{equation}
where $\tau_r$ and $\tau_m$ control the required prevalence and severity of cross-view inconsistency.
Tracks that produce no Gaussian support or no valid comparisons are also removed.
After filtering, we discard all masks belonging to the rejected objects, reinitialize the object-specific Gaussians from the remaining tracks, and train the final reconstruction.

\subsubsection{\textbf{Observation-level Filtering}}
An object retained by track-level filtering may still have erroneous masks or observations strongly affected by occlusion.
Because object-level supervision renders each Gaussian group independently, a heavily occluded mask may disagree with the projected object shape even when its visible region is correctly segmented.
We therefore select observations whose masks agree with the independently rendered object, allowing object-level regularization to rely on a subset of informative views.
This filtering is enabled during the later stage of final optimization, after the reconstruction has become sufficiently stable.
For each sampled object-view pair $(o,t)$, we independently render the object's alpha map $\hat{A}_t^o$ and obtain a binary mask
\begin{equation}
\bar{M}_t^o(u) =
\mathbb{I}\left[\hat{A}_t^o(u) > \tau_{\alpha}\right].
\end{equation}
We then define an observation-selection indicator as
\begin{equation}
w_t^o =
\mathbb{I}\left[
\mathrm{IoU}\left(\bar{M}_t^o,M_t^o\right)
\ge \tau_f
\right],
\end{equation}
where $\tau_f$ is the observation-level consistency threshold.
The object-level supervision for this pair becomes
\begin{equation}
\widetilde{\mathcal{L}}_{\mathrm{obj},t}^o
=
w_t^o
\left(
\mathcal{L}_{\mathrm{rgb},t}^o
+ \lambda_{\alpha}\mathcal{L}_{\alpha,t}^o
\right).
\end{equation}
When $w_t^o=0$, we skip the object RGB and alpha losses for the sampled pair $(o,t)$ while retaining the object track and full-scene photometric supervision.
The gate can exclude both erroneous masks and correctly segmented but heavily occluded observations.
Object-level regularization can thus rely on a subset of reliable views, while scene reconstruction continues to use the full set of training images.
Together, track-level filtering removes globally unreliable tracks, and observation-level filtering selects supervision compatible with independent object reconstruction.

\subsection{Viewpoint-aware Embedding Assignment}
\label{subsec:viewpoint_aware_embedding}
\begin{figure}
    \centering
    \includegraphics[width=1\linewidth]{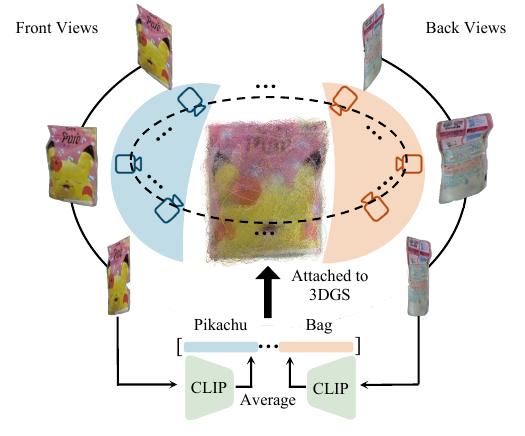}
    \caption{\textbf{Viewpoint-aware embedding assignment.}  Object crops from similar camera-pose clusters are encoded by CLIP and averaged within each cluster. The resulting cluster embeddings are attached to the same object-specific Gaussian group, preserving viewpoint-specific cues that a single global average can dilute.}
    \label{fig:viewpoint_descriptors}
\end{figure}
Unlike previous methods \cite{wu2024opengaussian, lu2026segment} that collapse all observations of an object into a single object-level embedding, we group object observations according to their camera viewpoints and aggregate CLIP features within each viewpoint cluster.
The intuition is that many objects exhibit view-dependent semantic cues.
For example, the packaging bag in \cref{fig:viewpoint_descriptors} bears a Pikachu graphic on its front, providing a distinctive cue for queries such as ``Pikachu'' that may be absent from other viewpoints.
A single globally averaged embedding can dilute this cue, making retrieval from such specific descriptions less discriminative.

We first group training views according to their camera poses. 
For each view $t$, we construct a pose feature by concatenating the camera center and viewing direction. 
The pose features are standardized and clustered into $K$ viewpoint groups using average-linkage hierarchical clustering, producing a global view-to-cluster assignment shared by all objects.
We compare this strategy with k-means and other aggregation strategies in the ablation studies.

For each object $o$, we collect its valid observations across the training views. 
Given image $I_t$ and the corresponding object mask $M_t^o$, we isolate the object region by masking out the background and cropping the result to the mask bounding box. 
Empty masks are skipped. 
Each valid crop is then encoded by the CLIP image encoder and $\ell_2$-normalized:
\begin{equation}
\mathbf{f}_{o,t} =
\frac{
\mathrm{CLIP}_{\mathrm{img}}(\mathrm{crop}(I_t, M_t^o))
}{
\left\|
\mathrm{CLIP}_{\mathrm{img}}(\mathrm{crop}(I_t, M_t^o))
\right\|_2
},
\end{equation}
where $\mathbf{f}_{o,t}$ denotes the visual-language feature of object $o$ observed from view $t$.

Instead of directly averaging features from all views, we aggregate them within each viewpoint cluster. 
For object $o$ and viewpoint cluster $c$, the cluster-level embedding is computed as
\begin{equation}
\mathbf{e}_{o,c} =
\frac{
\sum_{t \in \mathcal{V}_{o,c}} \mathbf{f}_{o,t}
}{
\left\|
\sum_{t \in \mathcal{V}_{o,c}} \mathbf{f}_{o,t}
\right\|_2
},
\end{equation}
where $\mathcal{V}_{o,c}$ denotes the set of valid views in which object $o$ is visible and whose camera pose belongs to cluster $c$. 
As a result, each object is represented by a set of viewpoint-aware language embeddings:
\begin{equation}
\mathcal{E}_o = \{\mathbf{e}_{o,c} \mid c \in \mathcal{C}_o\},
\end{equation}
where $\mathcal{C}_o$ is the set of viewpoint clusters containing valid observations of object $o$.

\subsection{Open-vocabulary Query}
\label{subsec:open_vocab_query}

Given the viewpoint-aware embeddings assigned to each object, we perform open-vocabulary querying at the object level. 
We pool the middle-level and large-level Gaussian groups into a joint candidate set $\mathcal{O}$, treating each group at each granularity as a distinct candidate.
For a text query $q$, we first extract its CLIP text embedding and normalize it as
\begin{equation}
\mathbf{e}_q =
\frac{
\mathrm{CLIP}_{\mathrm{text}}(q)
}{
\left\|
\mathrm{CLIP}_{\mathrm{text}}(q)
\right\|_2
}.
\end{equation}
For each object $o$, we compute its relevance score by comparing the text embedding with all available viewpoint-clustered embeddings:
\begin{equation}
s(o,q) =
\max_{c \in \mathcal{C}_o}
\cos\left(\mathbf{e}_{o,c}, \mathbf{e}_q\right),
\end{equation}
where $\mathcal{C}_o$ denotes the set of viewpoint clusters with valid observations for object $o$, and $\mathbf{e}_{o,c}$ is the corresponding cluster-level embedding.

We rank the candidates jointly across both granularities and select the group with the highest cosine-similarity score:
\begin{equation}
o^{*} = \arg\max_{o\in\mathcal{O}} s(o,q).
\end{equation}
Since each object is represented by an explicit set of Gaussian primitives, the query result can be directly obtained in 3D by extracting the corresponding object-specific Gaussian groups.

\section{Datasets \& Metrics}
\begin{figure*}
    \centering
    \includegraphics[width=1\linewidth]{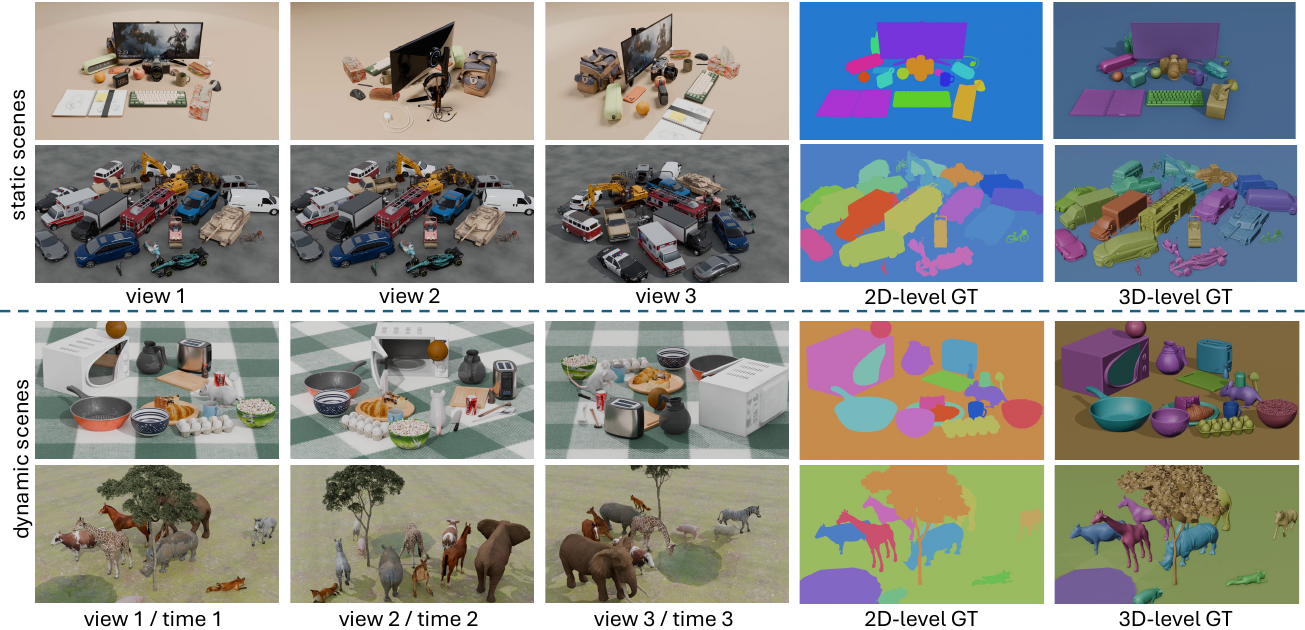}
    \caption{\textbf{Examples from Blender-OVS.} Static scenes (top) are rendered from multiple calibrated viewpoints; dynamic scenes (bottom) additionally vary object pose and visibility over time. Each sequence includes dense per-view instance masks (2D-level GT) and exact object geometry in world coordinates (3D-level GT), shown with consistent instance colors. The benchmark contains five static and five dynamic scenes.}
    \label{fig:blender_ovs}
\end{figure*}

\subsection{Datasets}
\begin{table*}[t]
\centering
\caption{Comparison of existing datasets for open-vocabulary segmentation.}
\label{tab:dataset_comparison}
\resizebox{0.9\textwidth}{!}{
\begin{tabular}{lccccc}
\toprule
\textbf{Dataset} & \textbf{\shortstack{Static Objects}} & \textbf{\shortstack{Dynamic Objects}} & \textbf{\shortstack{3D Instance GT}} & \textbf{\shortstack{Exact 2D-3D Alignment}} & \textbf{\shortstack{Paired Language}} \\
\midrule
3DOVS~\cite{liu2023weakly} & \textcolor{green!50!black}{$\checkmark$} & \textcolor{red}{$\times$} & \textcolor{red}{$\times$} & \textcolor{red}{$\times$} & \textcolor{green!50!black}{$\checkmark$} \\
LERF~\cite{kerr2023lerf} & \textcolor{green!50!black}{$\checkmark$} & \textcolor{red}{$\times$} & \textcolor{red}{$\times$} & \textcolor{red}{$\times$} & \textcolor{green!50!black}{$\checkmark$} \\
Mip-NeRF360~\cite{barron2022mip} & \textcolor{green!50!black}{$\checkmark$} & \textcolor{red}{$\times$} & \textcolor{red}{$\times$} & \textcolor{red}{$\times$} & \textcolor{green!50!black}{$\checkmark$} \\
ScanNet++~\cite{yeshwanth2023scannetpp} & \textcolor{green!50!black}{$\checkmark$} & \textcolor{red}{$\times$} & \textcolor{green!50!black}{$\checkmark$} & \textcolor{red}{$\times$} & \textcolor{green!50!black}{$\checkmark$} \\
HyperNeRF~\cite{park2021hypernerf} & \textcolor{red}{$\times$} & \textcolor{green!50!black}{$\checkmark$} & \textcolor{red}{$\times$} & \textcolor{red}{$\times$} & \textcolor{green!50!black}{$\checkmark$} \\
Neu3D~\cite{li2022neural3dvideo} & \textcolor{red}{$\times$} & \textcolor{green!50!black}{$\checkmark$} & \textcolor{red}{$\times$} & \textcolor{red}{$\times$} & \textcolor{green!50!black}{$\checkmark$} \\
\midrule
\textbf{Blender-OVS (ours)} & \textcolor{green!50!black}{$\checkmark$} & \textcolor{green!50!black}{$\checkmark$} & \textcolor{green!50!black}{$\checkmark$} & \textcolor{green!50!black}{$\checkmark$} & \textcolor{green!50!black}{$\checkmark$} \\
\bottomrule
\end{tabular}
}
\end{table*}

\subsubsection{\textbf{2D Pixel-level Open-vocabulary Segmentation}}
For static scenes, we evaluate rendered-view open-vocabulary segmentation on LERF~\cite{kerr2023lerf} and Mip-NeRF360~\cite{barron2022mip}. Following prior 3D open-vocabulary segmentation works~\cite{qin2024langsplat,shi2024language}, these benchmarks evaluate
segmentation quality in the 2D image space by rendering the queried 3D representation from selected camera views and comparing the resulting masks with pixel-level annotations.
For dynamic scenes, we evaluate on HyperNeRF~\cite{park2021hypernerf}, following the rendered-view open-vocabulary segmentation protocol of Segment-then-Splat~\cite{lu2026segment}.
This protocol measures whether a method can localize language-specified objects from observed viewpoints and provides a direct comparison
with existing language-embedded scene representations. However, since the evaluation results are defined on rendered images, these datasets do not
directly assess whether the underlying 3D object geometry is complete, consistently segmented across views, or correctly separated from neighboring objects.

\subsubsection{\textbf{3D Instance-level Open-vocabulary Segmentation}}
To evaluate open-vocabulary segmentation at the 3D object level, we use six scenes from ScanNet++~\cite{yeshwanth2023scannetpp}, which provides manually annotated semantic and instance labels on scan-derived indoor meshes. Additionally we introduce \textit{Blender-OVS}, a synthetic benchmark with complete and exact object-level geometric annotations.

While ScanNet++ enables 3D evaluation on real captures, the coverage and accuracy of its reference geometry depend on scanning and reconstruction, and it contains no dynamic objects (\cref{tab:dataset_comparison}).
\textit{Blender-OVS} fills both gaps: each scene provides calibrated camera intrinsics and extrinsics, object identities, appearance attributes and mesh geometry in the world coordinate system, all exported directly from the source assets.
The images, instance masks, and 3D geometry share the same rendering scene and camera parameters, providing \textit{exact 2D--3D alignment}.
This correspondence is defined directly by the renderer, without estimating alignment through reconstruction or scan registration.
Each object is paired with language tags linked to the same instance identity used for its rendered masks and source mesh, providing a consistent correspondence between language, image-space annotations, and 3D geometry.
These annotations support evaluation of both rendered-view segmentation quality and 3D object extraction quality against the corresponding ground truth.

\subsubsection{\textbf{Benchmark construction}}
We construct Blender-OVS by collecting diverse 3D assets from publicly available sources~\cite{objaverseXL,objaverse} and composing them into complex scenes using Blender~\cite{blender}. The benchmark contains 10 scenes in total: 5 static scenes for evaluating multi-view segmentation under fixed object layouts, and 5 dynamic scenes with object motion and changing visibility (\eg, objects entering or leaving the field of view and mutual occlusion).

\subsection{2D Pixel-level Metrics}
For rendered-view segmentation, we evaluate pixel-level mask quality using mean Intersection-over-Union (mIoU) and mean Boundary Intersection-over-Union (mBIoU). Given a language query, we render the predicted 3D segmentation into the target view and compare the resulting binary mask with the annotated 2D ground-truth mask. Let $\hat{M}_i$ and $M_i$ denote the predicted and ground-truth masks for the $i$-th evaluated query-view pair, and let $N$ denote the number of evaluated pairs. mIoU measures the average region overlap between predicted and ground-truth masks:
\begin{equation}
\mathrm{mIoU} =
\frac{1}{N}\sum_{i=1}^{N}
\frac{|\hat{M}_i \cap M_i|}{|\hat{M}_i \cup M_i|}.
\end{equation}
mBIoU computes IoU over boundary regions and therefore better reflects the accuracy of object contours:
\begin{equation}
\mathrm{mBIoU} =
\frac{1}{N}\sum_{i=1}^{N}
\frac{|\mathcal{B}(\hat{M}_i) \cap \mathcal{B}(M_i)|}
{|\mathcal{B}(\hat{M}_i) \cup \mathcal{B}(M_i)|},
\end{equation}
where $\mathcal{B}(\cdot)$ denotes the boundary region extracted from a binary mask. These two metrics jointly evaluate whether a method can localize the
queried object and preserve accurate object boundaries in the rendered image space.

\subsection{3D Instance-level Metrics}
For 3D object-level evaluation on \textit{Blender-OVS} and ScanNet++, we use normalized Chamfer distance to measure the discrepancy between the predicted and ground-truth object point sets.
For each query, the predicted point cloud $\mathcal{P}$ consists of the centers of the retrieved Gaussian primitives, and the ground-truth point cloud $\mathcal{Q}$ consists of the corresponding object mesh vertices.
We normalize the ground-truth object's axis-aligned bounding box to a unit box and apply the same translation and per-axis scaling factors to the predicted point cloud.
Using the normalized point clouds, we compute the bidirectional Chamfer distance based on L2 nearest-neighbor distances:
\begin{equation}
\begin{aligned}
d_{\mathrm{CD}}(\mathcal{P}, \mathcal{Q}) =
& \frac{1}{2|\mathcal{P}|}
\sum_{\mathbf{p}\in\mathcal{P}}
\min_{\mathbf{q}\in\mathcal{Q}}
\|\mathbf{p}-\mathbf{q}\|_2 \\
& +
\frac{1}{2|\mathcal{Q}|}
\sum_{\mathbf{q}\in\mathcal{Q}}
\min_{\mathbf{p}\in\mathcal{P}}
\|\mathbf{q}-\mathbf{p}\|_2 .
\end{aligned}
\end{equation}
A lower normalized Chamfer distance indicates closer agreement between the retrieved Gaussian centers and the ground-truth mesh vertices under this point-set evaluation protocol.

\subsection{Reconstruction Quality}
In both the 2D pixel-level and 3D instance-level experiments, we evaluate scene reconstruction quality using Peak Signal-to-Noise Ratio (PSNR)~\cite{huynh2008scope}, Structural Similarity Index Measure (SSIM)~\cite{wang2004image}, and
Learned Perceptual Image Patch Similarity (LPIPS)~\cite{zhang2018perceptual} between rendered novel views and ground-truth images. PSNR measures pixel-level reconstruction fidelity,
SSIM evaluates structural similarity, and LPIPS reflects perceptual image similarity in a learned feature space. Higher PSNR and SSIM indicate better
reconstruction quality, while lower LPIPS indicates better perceptual quality.

\section{Experiments}
As mentioned in the previous section, we conduct two types of experiments to evaluate our method: 2D pixel-level open-vocabulary segmentation and 3D instance-level open-vocabulary segmentation. 
The former (\cref{subsec:2DOVS}) focuses on rendered-view segmentation quality, where the predicted 3D representation is queried by language, rendered into selected camera views, and compared with pixel-level ground-truth masks. 
This setting evaluates whether our method can accurately localize language-specified objects from observed viewpoints and produce precise object boundaries in the image space. 
The latter (\cref{subsec:3DOVS}) focuses on object-level 3D geometry quality, where the extracted 3D Gaussians are directly compared with ground-truth 3D object geometry. 
This setting further examines whether our method can recover complete object geometry and correctly separate target objects from surrounding distractors in 3D space.

\begin{table*}[thbp]
\centering
\caption{Quantitative comparisons on 2D pixel-level open-vocabulary segmentation, on static scenes (Mip-NeRF360, LERF) and dynamic scenes (HyperNeRF). Best results are in \textbf{bold}, and second-best results are \underline{underlined}.}
\label{tab:2d_quantitative_results}
\resizebox{0.8\textwidth}{!}{
\begin{tabular}{ccccccc}
\toprule
\textbf{Dataset} & \textbf{Method} & \textbf{PSNR$\boldsymbol{\uparrow}$} & \textbf{SSIM$\boldsymbol{\uparrow}$} & \textbf{LPIPS$\boldsymbol{\downarrow}$} & \textbf{mIoU$\boldsymbol{\uparrow}$} & \textbf{mBIoU$\boldsymbol{\uparrow}$} \\
\midrule

\multirow{9}{*}{LERF~\cite{kerr2023lerf}}
& LEGaussians~\cite{shi2024language} & 24.775 & 0.874 & 0.198 & 26.465 & 20.768 \\
& LangSplat~\cite{qin2024langsplat} & 24.002 & 0.856 & 0.218 & 47.463 & 41.373 \\
& Gaussian-Grouping~\cite{gaussian_grouping} & 23.861 & 0.862 & 0.211 & 29.360 & 26.690 \\
& OpenGaussian~\cite{wu2024opengaussian} & 22.642 & 0.855 & 0.213 & 43.778 & 38.380 \\
& ObjectGS~\cite{zhu2025objectgs} & 23.074 & 0.856 & 0.203 & 24.203 & 22.815 \\
& Segment-then-Splat~\cite{lu2026segment} & \underline{25.227} & \underline{0.882} & \underline{0.196} & 52.098 & \underline{48.978} \\
& GAGS~\cite{peng2026gags} & 24.318 & 0.868 & 0.205 & \underline{53.968} & 48.500 \\
& ReLaGS~\cite{Xie_2026_CVPR} & 24.061 & 0.865 & 0.226 & 45.205 & 39.688 \\
\cmidrule(lr){2-7}
& Ours & \textbf{25.491} & \textbf{0.886} & \textbf{0.191} & \textbf{62.845} & \textbf{57.298} \\

\midrule

\multirow{9}{*}{Mip-NeRF360~\cite{barron2022mip}}
& LEGaussians~\cite{shi2024language} & 27.425 & 0.857 & 0.147 & 55.680 & 33.995 \\
& LangSplat~\cite{qin2024langsplat} & 27.762 & 0.849 & 0.144 & 59.060 & 48.872 \\
& Gaussian-Grouping~\cite{gaussian_grouping} & 27.668 & 0.842 & 0.201 & 47.792 & 39.377 \\
& OpenGaussian~\cite{wu2024opengaussian} & \textbf{28.868} & \underline{0.871} & 0.138 & 39.937 & 28.827 \\
& ObjectGS~\cite{zhu2025objectgs} & 28.295 & 0.867 & \textbf{0.134} & 43.417 & 35.692 \\
& Segment-then-Splat~\cite{lu2026segment} & 28.284 & 0.865 & 0.139 & 55.310 & 44.487 \\
& GAGS~\cite{peng2026gags} & \underline{28.644} & 0.866 & \underline{0.136} & 59.123 & \underline{49.907} \\
& ReLaGS~\cite{Xie_2026_CVPR} & 27.941 & 0.852 & 0.163 & \underline{60.502} & 49.155 \\
\cmidrule(lr){2-7}
& Ours & 28.431 & \textbf{0.872} & 0.140 & \textbf{62.840} & \textbf{51.140} \\
\midrule

\multirow{6}{*}{HyperNeRF~\cite{park2021hypernerf}}
& LangSplat~\cite{qin2024langsplat} & 23.914 & 0.866 & 0.179 & 63.215 & 50.483 \\
& ObjectGS~\cite{zhu2025objectgs} & 24.019 & 0.855 & 0.175 & 38.413 & 31.173 \\
& DGD~\cite{labe2024dgd} & 29.676 & 0.719 & 0.311 & 7.965 & 3.113 \\
& 4D LangSplat~\cite{li20254dlangsplat4dlanguage} & \textbf{30.490} & 0.888 & 0.137 & 59.749 & 52.163 \\
& Segment-then-Splat~\cite{lu2026segment} & 29.638 & \textbf{0.898} & \textbf{0.128} & \underline{69.480} & \underline{56.376} \\
\cmidrule(lr){2-7}
& Ours & \underline{30.392} & \underline{0.893} & \underline{0.134} & \textbf{72.507} & \textbf{61.371} \\
\bottomrule
\end{tabular}
}
\end{table*}

\subsection{Implementation Details}
\label{subsec:implementation}
For mask tracking, we check for new objects every $\Delta t=10$ frames and trigger discovery when the fraction of uncovered image pixels increases by more than $0.01$ relative to the current reference frame.
A new mask is accepted if it contains more than 5,000 pixels and more than 50\% of its area is not covered by existing tracks.
For conflict resolution, we use $\tau_{\mathrm{iou}}=0.1$ and $\tau_c=0.3$, ignoring overlaps smaller than 64 pixels.
The reliability weights are $\lambda_{\mathrm{IoU}}=0.65$, $\lambda_{\mathrm{shape}}=0.20$, and $\lambda_{\mathrm{motion}}=0.15$.
Area stability is computed using the most recent five non-empty mask areas.

We use original 3DGS~\cite{kerbl3Dgaussians} for static scenes and the HexPlane-based 4DGS backbone~\cite{Wu_2024_CVPR} for dynamic scenes.
To obtain reconstruction feedback for mask correction, we use intermediate checkpoints at 7,000 iterations for static scenes and 10,000 iterations for dynamic scenes.
After track-level filtering, we reinitialize the object-specific Gaussians using the retained masks and train the final reconstruction for 40,000 iterations.
In each dynamic reconstruction pass, the first 3,000 iterations form a canonical warm-up stage, during which deformation and HexPlane regularization are disabled.
We then jointly optimize the Gaussian attributes, feature planes, and deformation decoder, and enable $\mathcal{L}_{\mathrm{reg}}$.
The warm-up iterations are included in the reconstruction iteration counts.

We optimize Gaussian attributes using Adam and represent color with degree-three spherical harmonics.
We set $\lambda_r=0.2$, $\lambda_{\mathrm{obj}}=0.1$, and $\lambda_\alpha=1$.
For object-level supervision, we sample three object groups per active granularity in static scenes and one in dynamic scenes at each iteration.
Densification continues until iteration 20,000 of the final reconstruction.
For track-level filtering, we set $\tau_v=0.5$, $\tau_{\mathrm{bad}}=0.3$, $\tau_r=0.4$, and $\tau_m=0.5$, requiring at least five valid view comparisons for the ratio and median criteria.
A view comparison is excluded if either mask contains fewer than 64 pixels or if the ratio of the smaller mask area to the larger one is below $0.2$.
Observation-level filtering is enabled after iteration 30,000 of the final reconstruction, with $\tau_\alpha=0.5$ and $\tau_f=0.3$.

After reconstruction, we encode tight object crops with a frozen CLIP encoder, masking the background to black.
We standardize the concatenated camera-center and viewing-direction and apply hierarchical clustering, using $K=\max(1,\mathrm{round}(\sqrt{N_{\mathrm{train}}}))$ clusters shared by all objects, where $N_{\mathrm{train}}$ is the number of training views.
For quantitative evaluation, each query retrieves the highest-scoring Gaussian group across both granularities.
For BIoU evaluation, the boundary width is 2\% of the image diagonal.

\subsection{2D Pixel-level Open-vocabulary Segmentation}
\label{subsec:2DOVS}
\subsubsection{\textbf{Baselines}}
For static scenes, we compare our method with eight Gaussian-based open-vocabulary segmentation approaches: LEGaussians~\cite{shi2024language}, LangSplat~\cite{qin2024langsplat}, Gaussian-Grouping~\cite{gaussian_grouping}, OpenGaussian~\cite{wu2024opengaussian}, ObjectGS~\cite{zhu2025objectgs}, GAGS~\cite{peng2026gags}, ReLaGS~\cite{Xie_2026_CVPR}, and our previous Segment-then-Splat framework~\cite{lu2026segment}.
For dynamic scenes, we include LangSplat~\cite{qin2024langsplat} and ObjectGS~\cite{zhu2025objectgs} as representative static-scene baselines, together with the dynamic Gaussian methods DGD~\cite{labe2024dgd} and 4D LangSplat~\cite{li20254dlangsplat4dlanguage}, as well as Segment-then-Splat~\cite{lu2026segment}.

\subsubsection{\textbf{Quantitative Results}}
The quantitative comparison results are shown in \cref{tab:2d_quantitative_results}.
On the static LERF scenes, our method achieves the best performance across all rendering and segmentation metrics.
In particular, it improves mIoU from 53.968 to 62.845 and mBIoU from 48.978 to 57.298 over the strongest competing results, demonstrating more accurate localization and substantially cleaner object boundaries.
On Mip-NeRF360, comparing with the strongest baselines, it improves mIoU by 2.338 points over ReLaGS and mBIoU by 1.233 points over GAGS.

The advantage is maintained in dynamic scenes.
On HyperNeRF, our method achieves an mIoU of 72.507 and an mBIoU of 61.371, outperforming Segment-then-Splat by 3.027 and 4.995 points, respectively.
DGD's weak segmentation results may partly reflect the limited representational capacity of its LSeg-based~\cite{li2022language} language features for fine-grained open-vocabulary queries.

\subsubsection{\textbf{Qualitative Results}}
\begin{figure*}[htbp]
    \centering
    \includegraphics[width=0.9\linewidth]{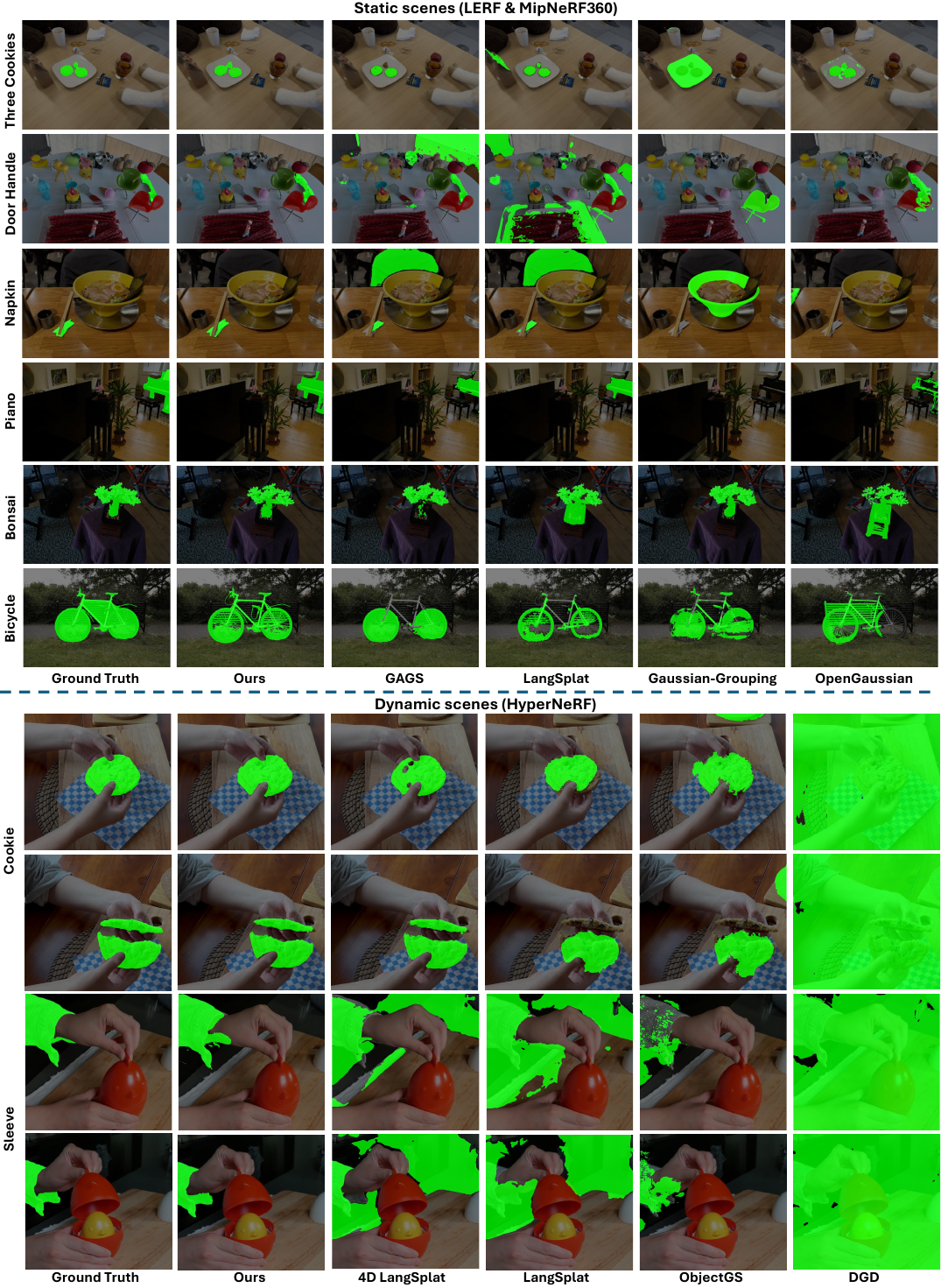}
    \caption{\textbf{Qualitative results for 2D pixel-level Open-vocabulary Segmentation}. Predicted segmentation is overlaid in green. The top block contains static LERF and Mip-NeRF360 queries, and the bottom block contains two timestamps each for HyperNeRF queries. RefineSplat more consistently suppresses neighboring objects and follows thin structures than the compared outputs in these examples.}
    \label{fig:2DOVS}
\end{figure*}

Qualitative results are shown in \cref{fig:2DOVS}.
Compared with the baselines, our method achieves more accurate object localization and produces cleaner, more precise boundaries, consistent with the quantitative improvements in mIoU and mBIoU.
We also observe that in some scenes, such as \textit{Bonsai} and \textit{Bicycle}, our predicted masks delineate thin structures (\eg, branches and spokes) at a finer level of detail than the available ground-truth annotations. This suggests that the reported IoU may underestimate the actual boundary quality in such cases.

\begin{table*}[t]
\centering
\caption{Quantitative comparisons on 3D instance-level open-vocabulary segmentation on ScanNet++ and Blender-OVS. For each dataset, best results are in \textbf{bold}, and second-best results are \underline{underlined}.}
\label{tab:3d_quantitative_results}
\resizebox{0.8\textwidth}{!}{
\begin{tabular}{cccccc}
\toprule
\textbf{Dataset} & \textbf{Method} & \textbf{PSNR$\boldsymbol{\uparrow}$} & \textbf{SSIM$\boldsymbol{\uparrow}$} & \textbf{LPIPS$\boldsymbol{\downarrow}$} & \textbf{Chamfer Distance$\boldsymbol{\downarrow}$} \\
\midrule

\multirow{5}{*}{ScanNet++~\cite{yeshwanth2023scannetpp}}
& Gaussian-Grouping~\cite{gaussian_grouping} & 22.412 & 0.789 & 0.463 & 2.128 \\
& OpenGaussian~\cite{wu2024opengaussian} & 23.153 & \underline{0.801} & 0.426 & \underline{1.605} \\
& ObjectGS~\cite{zhu2025objectgs} & \underline{23.213} & \textbf{0.821} & 0.403 & 3.498 \\
& Segment-then-Splat~\cite{lu2026segment} & 23.062 & 0.799 & \underline{0.348} & 1.766 \\
\cmidrule(lr){2-6}
& Ours & \textbf{23.278} & 0.800 & \textbf{0.344} &\textbf{1.284} \\

\midrule

\multirow{5}{*}{Blender-OVS (static)}
& Gaussian-Grouping~\cite{gaussian_grouping} & \underline{37.836} & \underline{0.969} & \underline{0.148} & 3.232 \\
& OpenGaussian~\cite{wu2024opengaussian} & 35.772 & 0.963 & 0.150 & \underline{2.021} \\
& ObjectGS~\cite{zhu2025objectgs} & 33.652 & 0.961 & 0.156 & 3.318 \\
& Segment-then-Splat~\cite{lu2026segment} & 37.764 & \underline{0.969} & 0.150 & 2.310 \\
\cmidrule(lr){2-6}
& Ours & \textbf{38.060} & \textbf{0.973} & \textbf{0.146} & \textbf{1.882} \\

\midrule

\multirow{6}{*}{Blender-OVS (dynamic)}
& Gaussian-Grouping~\cite{gaussian_grouping} & 32.166 & 0.934 & \underline{0.162} & 2.884 \\
& ObjectGS~\cite{zhu2025objectgs} & 30.335 & 0.928 & 0.169 & 3.231 \\
& DGD~\cite{labe2024dgd} & 32.263 & \underline{0.938} & 0.168 & 4.368 \\
& 4D LangSplat~\cite{li20254dlangsplat4dlanguage} & 29.886 & 0.846 & 0.316 & 3.503 \\
& Segment-then-Splat~\cite{lu2026segment} & \underline{32.279} & 0.935 & 0.167 & \underline{2.348} \\
\cmidrule(lr){2-6}
& Ours & \textbf{32.940} & \textbf{0.944} & \textbf{0.144} & \textbf{1.954} \\
\bottomrule
\end{tabular}
}
\end{table*}

\subsection{3D Instance-level Open-vocabulary Segmentation}
\label{subsec:3DOVS}
\subsubsection{\textbf{Baselines}}
For static scenes, we compare against Gaussian-Grouping~\cite{gaussian_grouping}, OpenGaussian~\cite{wu2024opengaussian}, ObjectGS~\cite{zhu2025objectgs}, and Segment-then-Splat~\cite{lu2026segment}.
For dynamic scenes, we compare against Gaussian-Grouping, ObjectGS, DGD~\cite{labe2024dgd}, 4D LangSplat~\cite{li20254dlangsplat4dlanguage}, and Segment-then-Splat.
DGD and 4D LangSplat learn Gaussian language features through feature-rendering supervision.
In our 3D evaluation of these methods, we directly retrieve Gaussian primitives by matching their language features to the text-query embedding, producing explicit object point sets for evaluation.

\subsubsection{\textbf{Quantitative Results}}
As shown in \cref{tab:3d_quantitative_results}, RefineSplat achieves the lowest Chamfer distance on all three benchmarks.
On ScanNet++, it reduces Chamfer distance from 1.605 to 1.284 relative to OpenGaussian, the strongest baseline on this metric, a reduction of 20.0\%.
On static Blender-OVS, RefineSplat obtains a Chamfer distance of 1.882, improving over OpenGaussian's 2.021 by 6.9\% and Segment-then-Splat's 2.310 by 18.5\%.

On dynamic Blender-OVS, RefineSplat reduces Chamfer distance from 2.348 for Segment-then-Splat to 1.954, a relative reduction of 16.8\%.
Together, these results indicate closer agreement between the retrieved object geometry and the target geometry in both static and dynamic scenes.
The metric evaluates the combined outcome of language-based object selection and geometric reconstruction.

\begin{figure*}[htbp]
    \centering
    \includegraphics[width=0.95\linewidth]{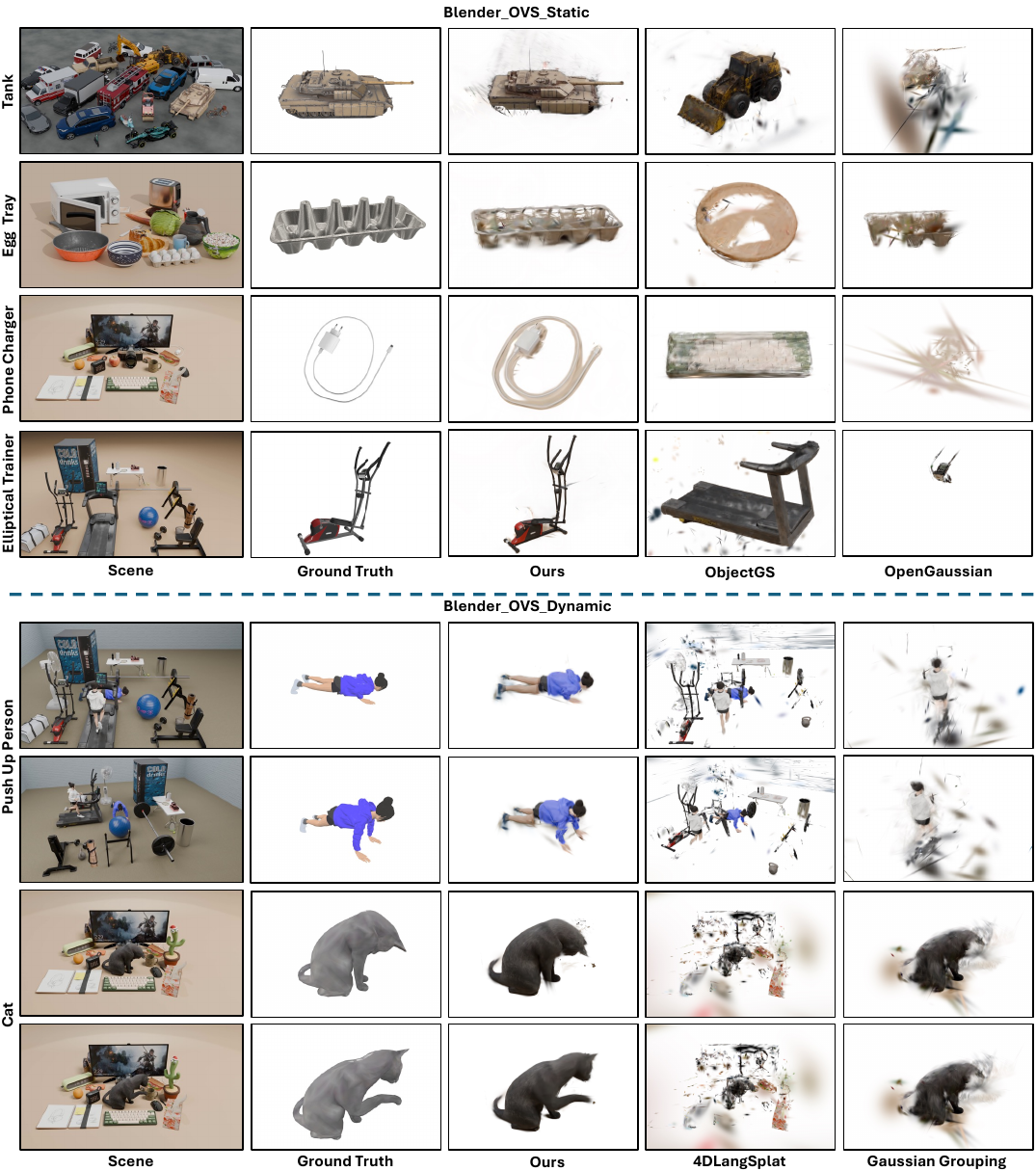}
    \caption{\textbf{Qualitative results for 3D instance-level Open-vocabulary Segmentation on Blender-OVS}. Each row shows the source scene, ground-truth object, and language-guided object extractions. The dynamic block (bottom) shows two timestamps for each query.}
    \label{fig:3DOVS}
\end{figure*}

Compared with HyperNeRF, the moving objects in dynamic Blender-OVS occupy a relatively small portion of the scene, as illustrated in \cref{fig:3DOVS}.
Full-scene PSNR, SSIM, and LPIPS are less sensitive to improvements on moving objects.
This helps explain the relatively modest reconstruction gains of dynamic methods over static baselines such as Gaussian-Grouping and ObjectGS.
In addition, 4D LangSplat reconstructs the \textit{kitchen} scene particularly poorly, lowering its average reconstruction scores across the dataset.

\subsubsection{\textbf{Qualitative Results}}
\Cref{fig:3DOVS} visualizes extracted objects on static and dynamic Blender-OVS.
In the static examples, ObjectGS retrieves semantically related but incorrect objects, such as a loader for \textit{Tank} and a treadmill for \textit{Elliptical Trainer}, while OpenGaussian produces incomplete extractions and scattered primitives in several queries.
RefineSplat retrieves the intended instances and better preserves their overall shapes, including the cable of \textit{Phone Charger} and the slender supports of \textit{Elliptical Trainer}.
For the dynamic \textit{Push Up Person} and \textit{Cat} queries, our extracted objects follow the target poses at both displayed timestamps.
For \textit{Push Up Person}, Gaussian-Grouping incorrectly retrieves the person on the treadmill.
The 4D LangSplat outputs contain many unrelated Gaussians around the queried objects, which may stem from its feature-rendering supervision~\cite{li20254dlangsplat4dlanguage}.
Alpha-blending constrains the aggregated feature along each camera ray, allowing individual Gaussian features to become entangled with semantic information from other objects.
Consequently, directly retrieving Gaussians based on these features can include substantial unrelated scene content.
Our outputs more clearly separate the queried objects from their surroundings.
These observations are consistent with the lower Chamfer distances in \cref{tab:3d_quantitative_results}.

\subsection{Ablation Studies}
\begin{table*}[t]
\centering
\caption{Ablation study of reconstruction-guided feedback modules.}
\label{tab:ablation_feedback}
\resizebox{0.85\textwidth}{!}{
\begin{tabular}{cccccccc}
\toprule
\multicolumn{3}{c}{\textbf{Module}} & \multicolumn{5}{c}{\textbf{Metric}} \\
\cmidrule(lr){1-3} \cmidrule(lr){4-8}
\textbf{SAM2} & \textbf{Track Filtering} & \textbf{Observation Filtering} & \textbf{PSNR$\boldsymbol{\uparrow}$} & \textbf{SSIM$\boldsymbol{\uparrow}$} & \textbf{LPIPS$\boldsymbol{\downarrow}$} & \textbf{mIoU$\boldsymbol{\uparrow}$} & \textbf{mBIoU$\boldsymbol{\uparrow}$} \\
\midrule
$\checkmark$ & - & - & 24.932 & 0.879 & 0.194 & 54.672 & 48.935 \\
$\checkmark$ & $\checkmark$ & - & 25.087 & 0.876 & 0.203 & 61.704 & 56.182 \\
$\checkmark$ & $\checkmark$ & $\checkmark$ & 25.491 & 0.886 & 0.191 & 62.845 & 57.298 \\
\bottomrule
\end{tabular}
}
\end{table*}

\subsubsection{\textbf{Reconstruction-guided Filtering}}
We ablate the track- and observation-level feedback mechanisms on LERF in \cref{tab:ablation_feedback}.
Using the original SAM2 masks without reconstruction-guided feedback provides the baseline performance.
Introducing track-level filtering improves mIoU from 54.672 to 61.704 and mBIoU from 48.935 to 56.182, corresponding to gains of 7.032 and 7.247 points, respectively.
These gains are consistent with the intended role of removing globally inconsistent tracks before final reconstruction, thereby reducing unreliable object-level supervision.
Adding observation-level filtering provides a further improvement to 62.845 mIoU and 57.298 mBIoU, while more clearly improving the reconstruction quality relative to track-level filtering.
The additional segmentation gains are 1.141 mIoU and 1.116 mBIoU, suggesting that the two filtering stages contribute differently: track-level filtering provides the main segmentation improvement, while observation-level filtering further refines the segmentation and improves appearance reconstruction.

\subsubsection{\textbf{Viewpoint-aware Embedding Assignment}}
\Cref{tab:ablation_query} compares different strategies for aggregating multi-view object embeddings on LERF.
The per-view strategy performs worst because individual observations can be sensitive to occlusion and viewpoint-specific ambiguity.
Averaging all views improves robustness but collapses distinct viewpoint-dependent cues into a single embedding.
Both view-clustering strategies alleviate this issue by preserving multiple representative embeddings for each object.
Hierarchical clustering achieves the best results, reaching 62.85 mIoU and 57.30 mBIoU.
It improves over multi-view averaging by 3.91 mIoU and 4.20 mBIoU, and over K-means clustering by 3.37 mIoU and 3.44 mBIoU.
We attribute the advantage of hierarchical clustering to its ability to adapt to non-uniform camera trajectories, whereas k-means tends to split dense pose regions and merge sparse ones.
\begin{table}[ht]
\centering
\caption{Ablation study of embedding assignment strategies.}
\label{tab:ablation_query}
\resizebox{0.9\columnwidth}{!}{
\begin{tabular}{lcc}
\toprule
\textbf{Query Strategy} & \textbf{mIoU$\boldsymbol{\uparrow}$} & \textbf{mBIoU$\boldsymbol{\uparrow}$} \\
\midrule
Global Multi-view Average & 58.94 & 53.10 \\
Independent Per-view & 56.60 & 51.24 \\
View Clusters (K-means) & 59.48 & 53.86 \\
View Clusters (Hierarchical) & \textbf{62.85} & \textbf{57.30}\\
\bottomrule
\end{tabular}
}
\end{table}

\subsection{Efficiency Analysis}
\label{subsec:efficiency}
\begin{table}[t]
\centering
\caption{Training and inference time analysis.}
\label{tab:efficiency}
\begingroup
\footnotesize
\begin{tabular}{@{}lcc@{}}
\toprule
\textbf{Method} & \textbf{\shortstack{Training time (h)$\boldsymbol{\downarrow}$}} & \textbf{\shortstack{Inference time (ms)$\boldsymbol{\downarrow}$}} \\
\midrule
\multicolumn{3}{c}{\textbf{Static Methods}} \\
\midrule
LEGaussians~\cite{shi2024language} & $\sim$2h & $\sim$1700ms \\
LangSplat~\cite{qin2024langsplat} & $\sim$2h & $\sim$100ms \\
Gaussian-Grouping~\cite{gaussian_grouping} & $\sim$1h & $\sim$700ms \\
OpenGaussian~\cite{wu2024opengaussian} & $\sim$1h & $\sim$10ms \\
ObjectGS~\cite{zhu2025objectgs} & $\sim$1h & $\sim$700ms \\
GAGS~\cite{peng2026gags} & $\sim$1h & $\sim$100ms \\
Segment-then-Splat~\cite{lu2026segment} & $\sim$1h & $\sim$10ms \\
\cmidrule(lr){1-3}
\textbf{RefineSplat (Ours)} & $\sim$1.5h & $\sim$10ms \\
\midrule
\multicolumn{3}{c}{\textbf{Dynamic Methods}} \\
\midrule
DGD~\cite{labe2024dgd} & $\sim$24h & $\sim$400ms \\
4D LangSplat~\cite{li20254dlangsplat4dlanguage} & $\sim$1.5h & $\sim$200ms \\
Segment-then-Splat~\cite{lu2026segment} & $\sim$3.5h & $\sim$10ms \\
\cmidrule(lr){1-3}
\textbf{RefineSplat (Ours)} & $\sim$2h & $\sim$10ms \\
\bottomrule
\end{tabular}
\endgroup
\end{table}

\Cref{tab:efficiency} presents the training- and inference-time comparison for the static and dynamic settings, using the baselines evaluated in \cref{subsec:2DOVS}.
RefineSplat requires approximately 1.5 hours per static scene.
For dynamic scenes, it takes approximately 2 hours per scene, compared with 3.5 hours for Segment-then-Splat.
Our reconstruction-guided refinement introduces additional computation through intermediate reconstruction and supervision filtering.
The lower overall training time in dynamic scenes results from replacing the Deformable 3DGS backbone~\cite{yang2024deformable} used by Segment-then-Splat with the more efficient 4DGS backbone~\cite{Wu_2024_CVPR}.
These additional operations improve supervision reliability while keeping the overall training time comparable to existing Gaussian-based pipelines.

RefineSplat also maintains fast inference, requiring approximately 10\,ms per query in both static and dynamic scenes, matching OpenGaussian and Segment-then-Splat in their respective settings.
Several other static methods require approximately 100-1700\,ms per query, while 4D LangSplat and DGD require approximately 200\,ms and 400\,ms, respectively, in the dynamic setting.
The shared efficiency of OpenGaussian, Segment-then-Splat, and RefineSplat comes from grouping Gaussians and associating each group with language embeddings before inference.
After encoding the text query, object selection requires only cosine-similarity comparisons against the stored group embeddings and retrieval of the corresponding Gaussians.
Other baselines require feature-map rendering and decoding~\cite{shi2024language,qin2024langsplat,li20254dlangsplat4dlanguage,labe2024dgd,peng2026gags}, or use Grounded-SAM~\cite{ren2024grounded} to select query-specific mask regions and associate them with Gaussian groups~\cite{gaussian_grouping,zhu2025objectgs}.
These additional feature-processing or 2D segmentation steps increase query latency.

\section{Limitations and Discussion}
\label{sec:limitations}

\noindent\textbf{Relational queries.}
RefineSplat associates language embeddings with individual Gaussian groups and matches queries to objects independently.
It therefore does not explicitly model relations between objects and cannot reliably resolve queries such as ``the cup to the left of the plate.''
Supporting such queries would require reasoning over the spatial or semantic relationships among multiple objects.

\noindent\textbf{Computational overhead.}
Reconstruction-guided feedback requires intermediate reconstruction and additional rendering and mask-consistency checks, followed by reinitialization for final reconstruction.
These steps introduce extra computational overhead during scene training compared with optimizing the final representation alone.
Although this overhead does not enter inference, reducing the cost of supervision refinement remains an important direction for future work.

\section{Conclusion}
We presented \textit{RefineSplat}, a reconstruction-guided object-centric framework for open-vocabulary segmentation and retrieval.
By decomposing a scene before reconstruction and optimizing dedicated Gaussian sets for individual objects, our segment-then-splat formulation preserves explicit object-Gaussian correspondence and enables queried objects to be directly retrieved as 3D Gaussian groups.
To reduce the impact of noisy tracking supervision, we introduced a reconstruction-guided feedback mechanism that uses intermediate object reconstructions to identify unreliable tracks and masks.
We further proposed a viewpoint-aware embedding assignment strategy that maintains multiple view-clustered embeddings for each object, avoiding the loss of discriminative semantic cues caused by averaging heterogeneous observations.

We also introduced \textit{Blender-OVS}, a benchmark with object-level mesh and point-cloud annotations for directly evaluating open-vocabulary object extraction in 3D, complementing conventional rendered-view evaluation.
Across the evaluated static and dynamic benchmarks, \textit{RefineSplat} improves rendered-view segmentation and achieves lower Chamfer distances for queried 3D objects.
Scene reconstruction quality remains competitive with the compared Gaussian-based methods.

\bibliographystyle{IEEEtran}
\bibliography{main}

\end{document}